%% file: main.tex
\documentclass[sigconf,nonacm]{acmart}
\usepackage{framed}
\usepackage[frozencache,cachedir=minted]{minted}

\AtBeginDocument{%
  \providecommand\BibTeX{{%
    \normalfont B\kern-0.5em{\scshape i\kern-0.25em b}\kern-0.8em\TeX}}}

\begin{document}

\title[DeeperDive: The Unreasonable Effectiveness of Weak Supervision in Document Understanding]{DeeperDive: The Unreasonable Effectiveness of Weak Supervision in Document Understanding\\
    \large A Case Study in Collaboration with UiPath Inc.}

\author{Emad Elwany}
\affiliation{%
  \institution{Lexion}
  \city{Seattle}
  \country{USA}
}
\email{emad@lexion.ai}

\author{Allison Hegel}
\affiliation{%
  \institution{Lexion}
  \city{Seattle}
  \country{USA}
}
\email{allison@lexion.ai}

\author{Marina Shah}
\affiliation{%
  \institution{Lexion}
  \city{Seattle}
  \country{USA}
}
\email{marina@lexion.ai}

\author{Brendan Roof}
\affiliation{%
  \institution{Lexion}
  \city{Seattle}
  \country{USA}
}
\email{brendan@lexion.ai}

\author{Genevieve Peaslee}
\affiliation{%
  \institution{Lexion}
  \city{Seattle}
  \country{USA}
}
\email{genevieve@lexion.ai}

\author{Quentin Rivet}
\affiliation{%
  \institution{Lexion}
  \city{Seattle}
  \country{USA}
}
\email{quentin@lexion.ai}

\renewcommand{\shortauthors}{Elwany et al.}

\begin{abstract}
Weak supervision has been applied to various Natural Language Understanding tasks in recent years. Due to technical challenges with scaling weak supervision to work on long-form documents, spanning up to hundreds of pages, applications in the document understanding space have been limited. At Lexion, we built a weak supervision-based system tailored for long-form (10-200 pages long) PDF documents. We use this platform for building dozens of language understanding models and have applied it successfully to various domains, from commercial agreements to corporate formation documents.

In this paper, we demonstrate the effectiveness of supervised learning with weak supervision in a situation with limited time, workforce, and training data. We built 8 high quality machine learning models in the span of one week, with the help of a small team of just 3 annotators working with a dataset of under 300 documents. We share some details about our overall architecture, how we utilize weak supervision, and what results we are able to achieve. We also include the dataset for researchers who would like to experiment with alternate approaches or refine ours.

Furthermore, we shed some light on the additional complexities that arise when working with poorly scanned long-form documents in PDF format, and some of the techniques that help us achieve state-of-the-art performance on such data.
\end{abstract}




\maketitle
\pagestyle{empty}

\input{background}
\input{dataset}
\input{approach}
\input{results}
\input{scalable_weak_supervision}
\input{conclusion}
\input{acknowledgments}

\bibliographystyle{ACM-Reference-Format}
\bibliography{main}

%\appendix

%\section{Appendix 1}

\end{document}

%% file: background.tex
\section{Background}

%At its heart, Lexion is a Document Intelligence company. As such we often get approached by partners who have interesting document understanding problems. REMOVED - SOUNDS TOO MUCH LIKE AN AD?
At Lexion, we are often approached by partners who rely on our Document Intelligence expertise to solve difficult document understanding problems.
UiPath, a leading enterprise automation software company, approached us in February 2022 with an exploratory project to evaluate the effectiveness of our platform. UiPath tasked us with extracting 8 key concepts from a set of 257 legal documents. These documents are ``Collective Bargaining Agreements” that govern the relationship between labor unions and employers.

\begin{figure}[t]
  \centering
  \includegraphics[width=\linewidth]{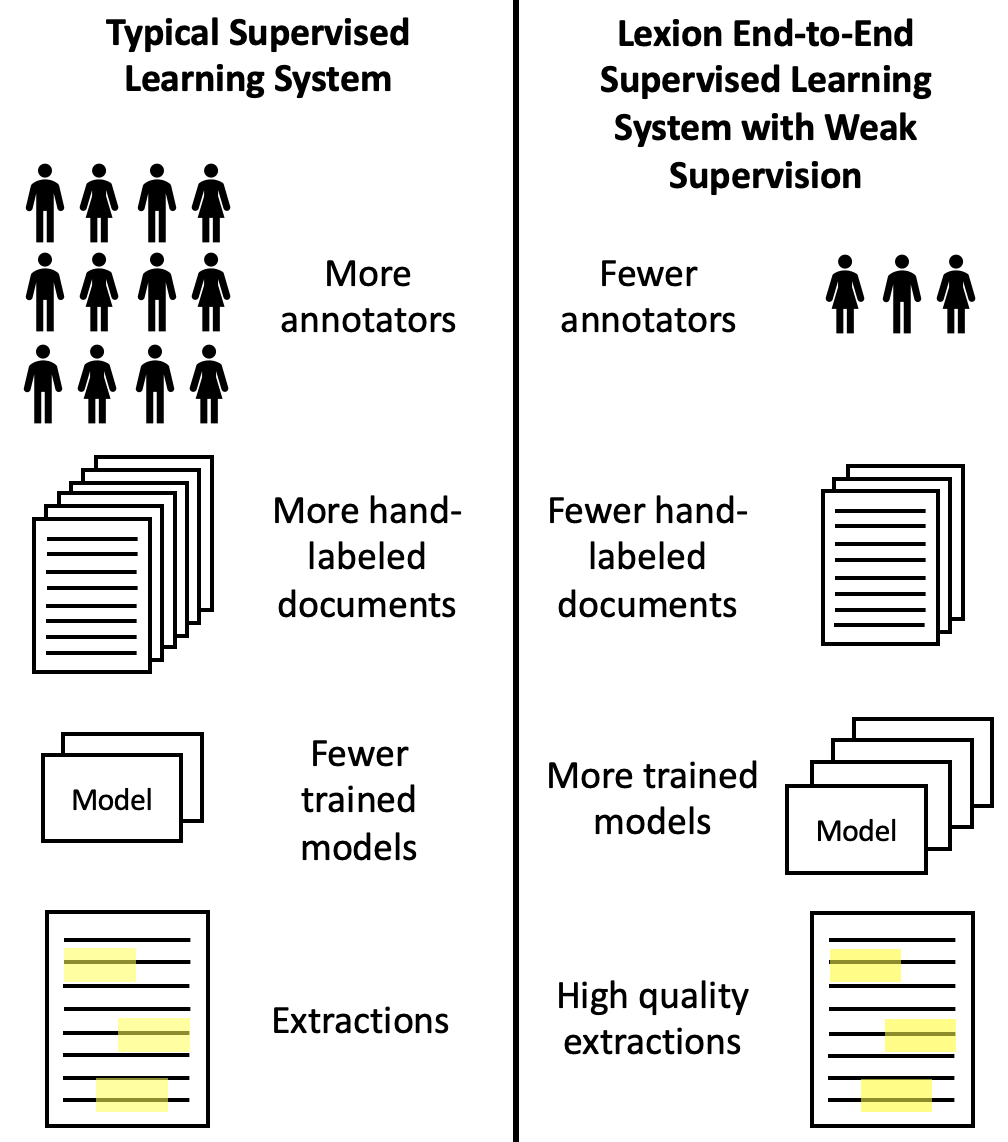}
  \caption{We develop an end-to-end system using weak supervision and deep neural networks to extract information from legal contracts with accuracy that rivals typical supervised learning systems.}
  \label{fig:frontpage}
\end{figure}

The goal was to see if we could build models to extract these 8 key concepts at high precision and recall with only one week of development. In addition to the time constraint, the dataset itself was particularly challenging.
The concepts appear throughout the documents in an unstructured manner, and these documents were 50 to 100 pages long on average. They were poorly scanned PDF documents with no native text information. The target values can show up in tables or in prose. They are regularly spread out over paragraphs  and are often interrupted by page boundaries, headers, footers and other elements, requiring a deep semantic understanding of the language to extract the correct values. We demonstrate that it is possible to achieve high precision and recall on this task despite constraints on time, workforce, and training data using an end-to-end supervised learning system that leverages weak supervision and deep neural models (\autoref{fig:frontpage}).

%% file: dataset.tex
\section{Dataset}

\begin{figure}[h]
  \centering
  \includegraphics[width=\linewidth]{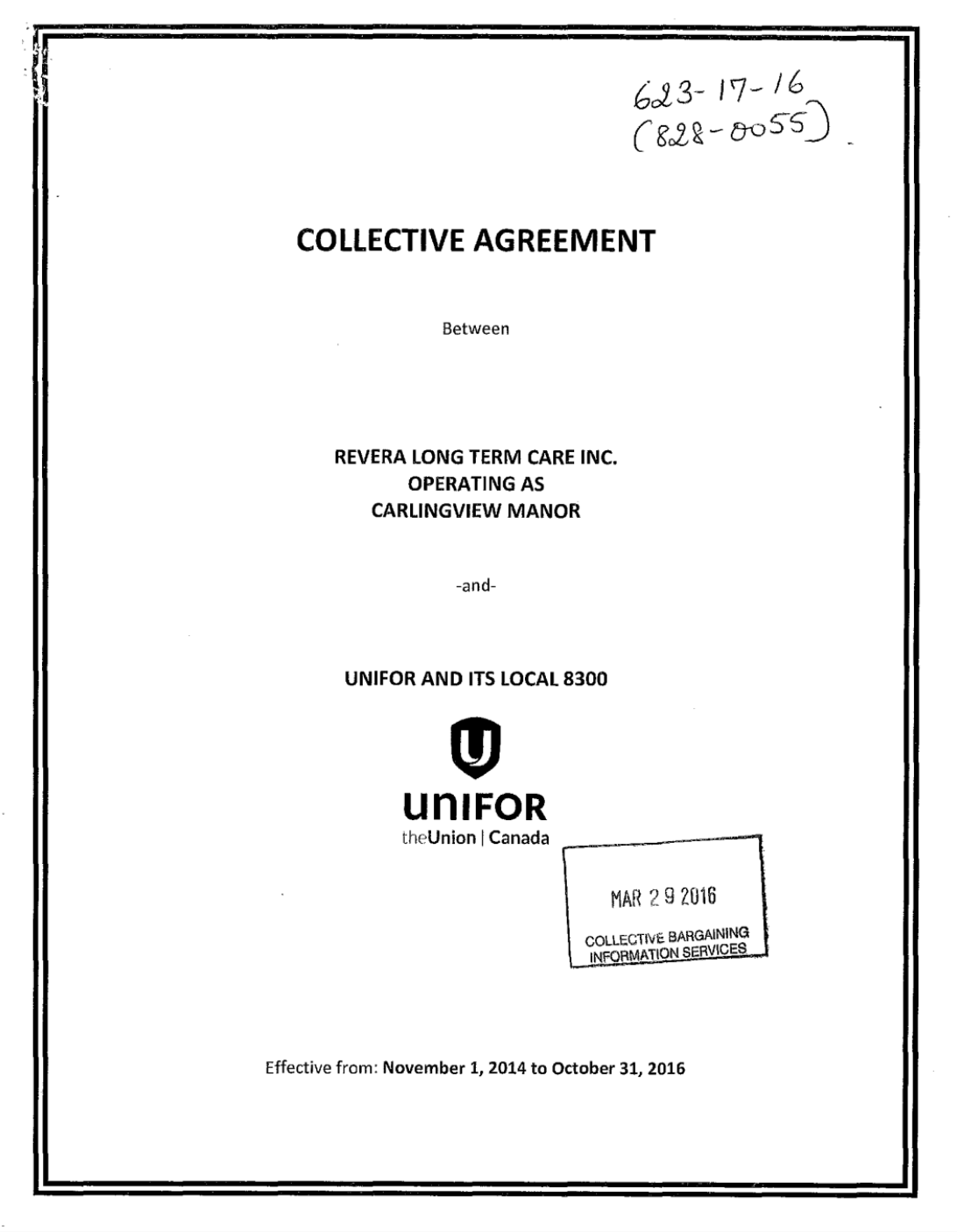}
  \caption{Cover page for a Collective Bargaining Agreement. Many documents in the dataset pose challenges such as poor scan quality, handwriting, tables, images, and stamps.}
    \label{fig:collective-agreement}
\end{figure}

\begin{figure}[h]
  \centering
  \includegraphics[width=\linewidth]{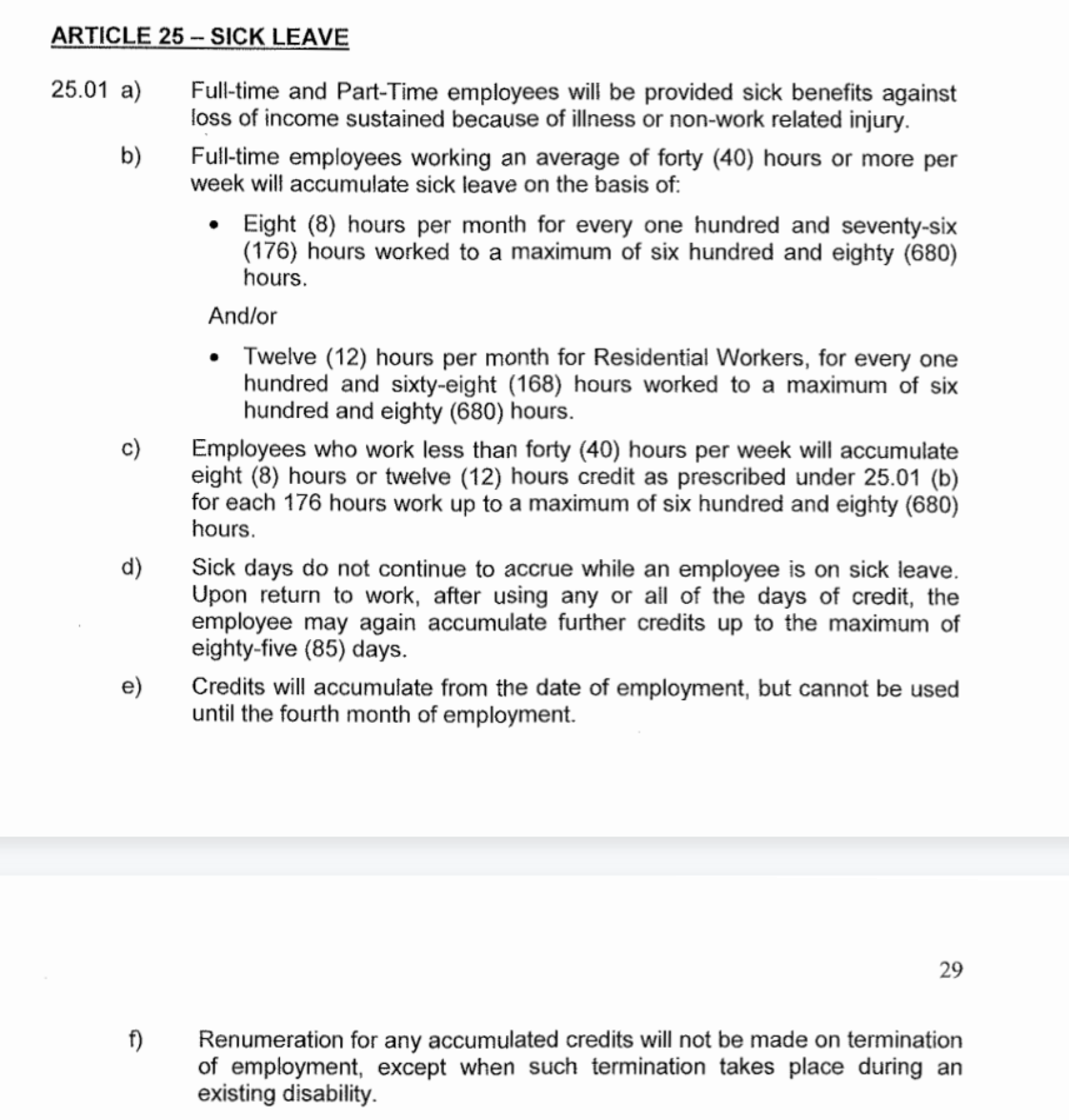}
  \caption{An example of a Sick Leave Clause from a Collective Bargaining Agreement in the dataset. These agreements pose challenges including section and subsection hierarchy, headers and footers, and page breaks.}
    \label{fig:sick-leave}
\end{figure}

Collective Bargaining Agreements are written contracts between an employer and a union representing the employees. The data used for this project consisted of 257 documents from the last 2 decades. Most of the documents were poorly scanned, and had complex layouts, obstructed text, and handwriting. \autoref{fig:collective-agreement} and \autoref{fig:sick-leave} show some representative examples of the data.

We extracted the following values from each agreement:

\begin{itemize}
  \item \textbf{Employer Name:} The name of the employer/company that is party to the contract.
  \item \textbf{Union Name:} The name of the labor union that is party to the contract.
  \item \textbf{Agreement Start Date:} The beginning of the fixed term for the contract.
  \item \textbf{Agreement End Date:} The end of the fixed term for the contract.
  \item \textbf{Sick Leave Clause:} The entire section and/or subsection discussing sick leave.
  \item \textbf{Sick Leave Amount per Employment Status:} The number of hours, shifts, days, or other unit of sick leave provided to employees.
  \item \textbf{Sick Leave Unit per Employment Status:} The unit listed after amount of sick leave, which may be per some amount of time worked (e.g. 8 hours per 2 weeks worked).
  \item \textbf{Employment Status:} The type of employee being referred to (e.g. part-time, full-time, or all employees).
\end{itemize}

Upon receiving the data, we divided it into a train set (80\% of the data), a development set (10\%), and a test set (10\%). We then ingested the data into our Document Intelligence platform. The development and test sets were manually annotated with perfect precision by our annotation team. Each document was first labeled by an annotator and then reviewed by a second annotator. The train set was annotated purely using weak supervision. Since long documents, particularly legal documents, are a particular challenge for natural language processing \citep{lexion1}, we have published our dataset for future research.\footnote{\url{https://drive.google.com/drive/u/0/folders/1rhglEd_IedBTJAF9G1KwdO2jii55FoeZ}}

%% file: approach.tex
\section{Approach}

\subsection{System Overview}

\begin{figure*}[h]
  \centering
  \includegraphics[width=\linewidth]{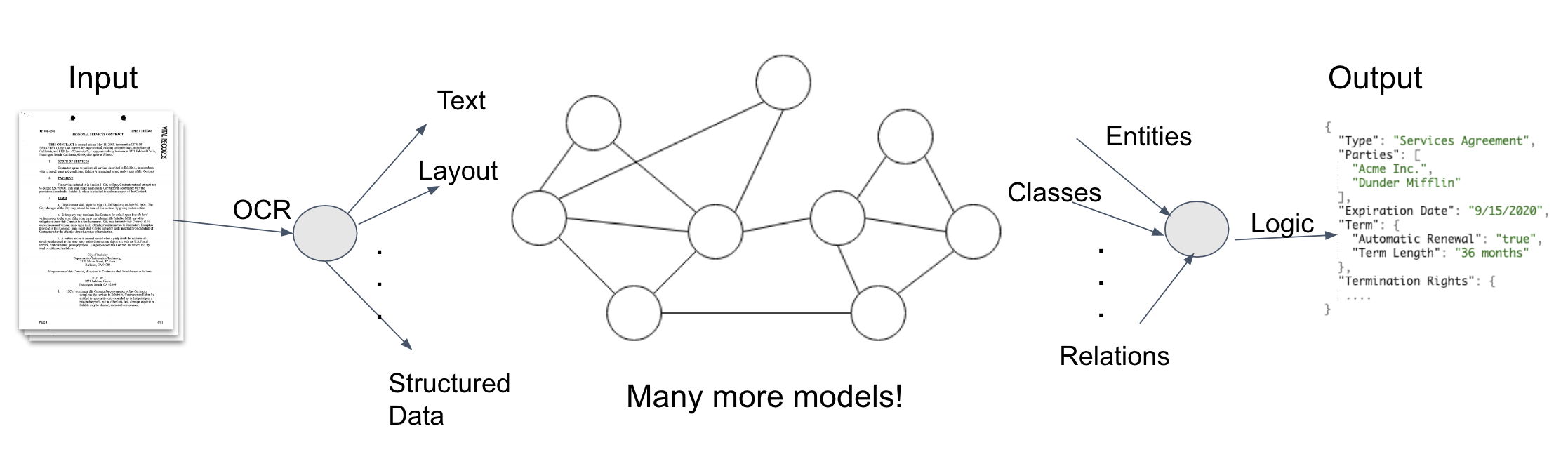}
  \caption{The Lexion Document Understanding Pipeline.}
    \label{fig:pipeline}
\end{figure*}

Lexion’s Document Understanding pipeline is comprised of multiple stages outlined in \autoref{fig:pipeline}. For this exercise, we leveraged many of our existing pre-trained models for the preliminary stages of the pipeline such as splitting PDFs that contain multiple agreements and detecting clauses. We focused primarily on training new models for the entity extraction and classification steps of the pipeline.

\subsection{Weak Supervision}

Our platform allows writing labeling functions using a domain-specific language. For an example of the syntax used by these labeling functions, see \autoref{tab:labeling-function1}.

\begin{table}[h]
\begin{framed}
\begin{minted}{python}
def label_sick_leave_hours(text) -> (start, end):
  for each sentence s in text:
    if s.starts("full time" or "part time") and
      s.contains("accumulate" or "accrue") and
      there exist tokens t1, t2 such that:
        if index(t1) - start(s) <= 5:
          index(t2) = index(t1) +1 and
          POS_TAG(t1) == "number" and
          NER_TAG(t2) == "time unit":
            return offsets(t1)
\end{minted}
\end{framed}
\caption{An example of a labeling function. The domain-specific language allows annotators to create functions that are highly specific and yet flexible enough to achieve high recall with only a limited number of functions.}
\label{tab:labeling-function1}
\end{table}

\begin{comment}
\begin{table}[h]
    \centering
    \footnotesize
    \begin{framed}
    \begin{tabular}{l}
        employee\_status: ((``all|full"?|``part"?)[]{0,1} ``time"? ``employees?") []{0,15}\\
        ``shall|will" ``accumu.*|accru.*" []{0,20} sick\_leave\_hours:([pos=``NUM"]{1,5})\\
        ``days?|hours?|shifts?" []{0,8} ``each|of|maximum|per|paid|period" []{0,20}\\
        ``maximum"
    \end{tabular}
    \end{framed}
    \caption{An example of labeling function syntax. The domain-specific language allows annotators to create functions that are highly specific and yet flexible enough to achieve high recall with only a limited number of functions.}
    \label{tab:labeling-function}
\end{table}
\end{comment}

By writing between 10 and 20 labeling functions for each concept that needed to be extracted, the annotation team was able to achieve 87-100\% annotation coverage for the training set in only a few days, with most models between 97-99\% coverage. We then used the platform to train the target models, which produces both train set and development set metrics. The trained models use neural network entity extraction and classification models that utilize large transformer language models with task-specific layers on top \citep{transformer,bert}. 

The annotation was completed by a team of 3 annotators over the span of 5 days. This time included not only writing labeling functions and annotating the test and development sets, but also doing schema and data discovery to understand the domain and gain the knowledge that was distilled into labeling functions. 

\subsection{Training Platform}

\begin{figure}[h]
  \centering
  \includegraphics[width=\linewidth]{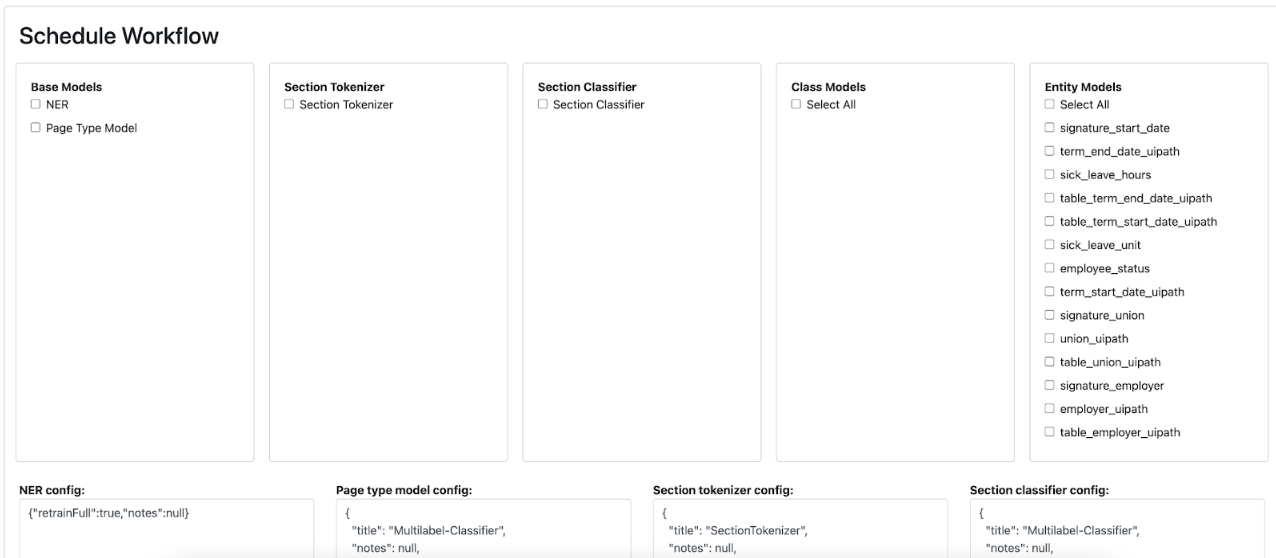}
  \caption{Our training platform allows the team to rapidly train models individually or as an end-to-end pipeline, while specifying model architecture and hyperparameters.}
  \label{fig:training-workflow}
\end{figure}

The training platform allows us to train the full pipeline or specific nodes very quickly by picking which nodes of the pipeline to train, and specifying the configurations we’d like, including model architecture and hyperparameters (\autoref{fig:training-workflow}). 

\begin{figure}[h]
  \centering
  \includegraphics[width=\linewidth]{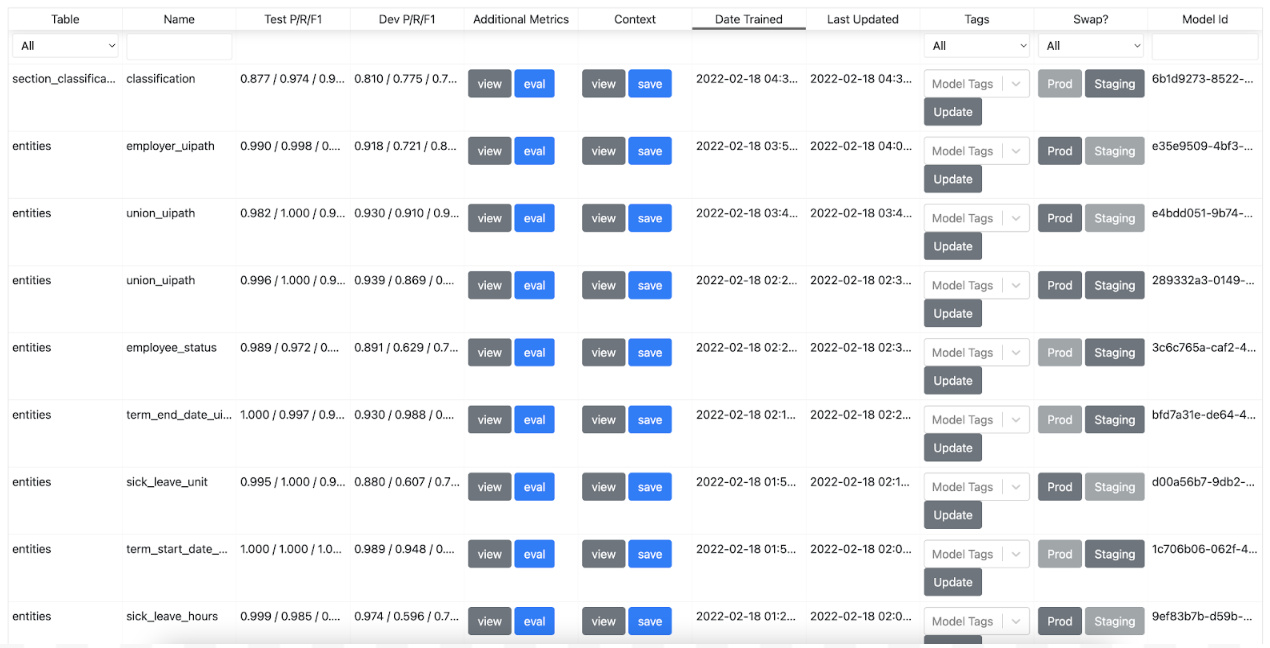}
  \caption{Our training platform offers detailed metrics on each model and allows the team to deploy newly-trained models with one click.}
  \label{fig:metric-deployment}
\end{figure}

In addition, the platform has metrics and deployment machinery built in (\autoref{fig:metric-deployment}), which allows us to quickly review the performance of individual models and roll them out to our user-facing Lexion interface.

\subsection{Lexion Interface}

\begin{figure}[h]
  \centering
  \includegraphics[width=\linewidth]{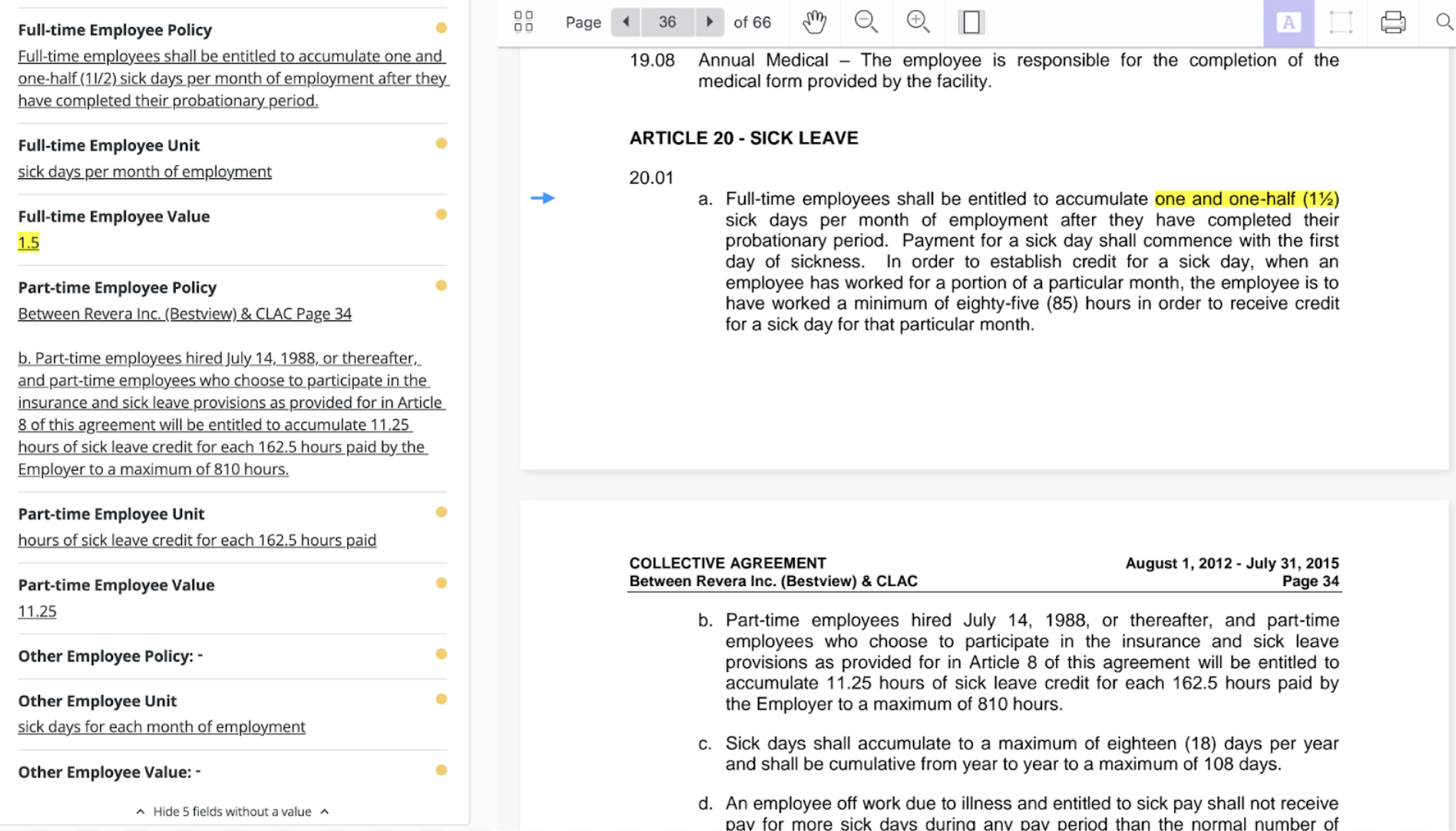}
  \caption{The Lexion interface displays the models' results in a user-friendly format, and allows users to correct any errors.}
  \label{fig:lexion-interface}
\end{figure}

We have found that a powerful user interface is an instrumental part of a document intelligence system. For that purpose, we have developed the Lexion interface (\autoref{fig:lexion-interface}) which makes it easy for non-technical users to view the results of document understanding extractions. 

This interface doesn’t just surface extractions to the end user, but also highlights the relevant language to provide explainability and confidence. It also allows users to verify the accuracy of extractions and correct mistakes, which is a powerful tool for closing the feedback loop between the user and the models. The user feedback we collect also allows further fine-tuning of the models.

%% file: results.tex
\section{Results}

With such a small and diverse dataset, by applying weak supervision for rapid annotation and a powerful neural network architecture, we were able to achieve impressive accuracy on the test set. We only evaluated the test set after training all the models, ensuring that it serves as a true blind set and was not subject to overfitting. \autoref{tab:results} demonstrates the results we were able to achieve on each desired concept.

\begin{table}[h]
\centering
\footnotesize
\begin{tabular}
{l | c c c | c c c }
Concept & Dev P & Dev R & Dev F1 & Test P & Test R & Test F1 \\
\hline
Employer Name       & 88.9 & 79.0 & 83.7 & 93.0 & 81.0 & 86.6\\
Union Name          & 94.1 & 91.0 & 92.5 & 96.8 & 80.1 & 87.7\\
Start Date          & 98.9 & 94.8 & 96.8 & 93.0 & 95.5 & 94.2\\
End Date            & 93.0 & 98.8 & 95.8 & 91.6 & 96.4 & 94.0\\
Sick Leave Clause   & 74.0 & 78.0 & 76.0 & 85.0 & 73.0 & 78.0\\
Sick Leave Amount   & 97.5 & 65.7 & 78.0 & 90.3 & 79.3 & 84.4\\
Sick Leave Unit     & 89.6 & 71.5 & 79.4 & 81.0 & 77.1 & 79.0\\
Employment Status   & 89.1 & 62.9 & 73.3 & 78.1 & 70.3 & 73.8\\
\hline
\end{tabular}
\caption{\label{tab:results}
\textbf{Precision, recall, and F1 score results on development and test data.}
}
\vspace{-1em}
\end{table}

Our models performed best on the two date-based concepts, Start Date and End Date, likely because of the strong named entity support of the domain-specific language we use to create labeling functions.
However, Employment Status was particularly challenging for the model because the contracts often discussed multiple statuses (e.g. both full-time and part-time) in the same clause.

The core differentiation of our system is the speed of model development. When the same techniques demonstrated in this paper are applied to larger datasets, for example the datasets that we use to train document understanding models on commercial agreements, the accuracy increases rapidly and often reaches F1 scores in the 0.85-0.95 range. On these large datasets (on the order of hundreds of thousands of documents), weak supervision is even more impactful since it would be prohibitively expensive to scale up model building with large amounts of annotation work.

%% file: scalable_weak_supervision.tex
\section{Scalable Weak Supervision}

%There has been successful application of weak supervision on short texts \citep{deepdive}, but much less work on longer texts. One of the main challenges of applying weak supervision effectively to long-form documents is building a robust and interactive system that can quickly evaluate labeling functions on huge volumes of documents \citep{lexion2}.
Typical supervised learning approaches require large amounts of labeled data. Weak supervision provides a method for generating labeled data at scale, while maintaining high accuracy.
Most applications of weak supervision focus on shorter texts like question-answer pairs and sentences from scientific journal articles \citep{deepdive,ratner,wang,lee}.
One of the main challenges of applying weak supervision effectively to long-form documents like legal contracts is building a robust and interactive system that can quickly evaluate labeling functions on large volumes of documents \citep{lexion2}.

%Not only do we have to operate on a large volume of long documents, but the documents are enriched with a lot of metadata that are used extensively in crafting labeling functions with high information density. In order to achieve this, we had to make large investments into our data platform, and had to deviate from the status quo of using Python as the language for writing labeling functions in favor of more efficient domain-specific languages offered by some open source frameworks. As a future goal, we hope to support labeling functions that have the full expressability of Python.
%We’d love to have the full expressability of Python, and getting Python labeling functions to work is a future goal for the team.
Particularly for longer documents, incorporating metadata into labeling functions has been shown to improve performance \citep{mekala}. Our approach enriches documents with metadata that is used extensively in crafting labeling functions with high information density. In order to achieve this, we have made large investments into our data platform, as well as deviating from the status quo of using Python as the language for writing labeling functions in favor of more efficient domain-specific languages offered by open source frameworks. As a future goal, we hope to support labeling functions that have the full expressability of Python.

%% file: conclusion.tex
\section{Conclusion}

%There has been a lot of recent work in document understanding.
With the introduction of large language models, multimodal architectures that employ both natural language processing and computer vision as well as deep neural networks, there have been great advances in the accuracy of models that convert unstructured text into structured data. The cost and time of annotation remains one of the main obstacles to scaling document understanding. 
We demonstrate in this paper an end-to-end application of weak supervision that achieves high performance at scale despite requiring less time, training data, and annotation resources than typical supervised learning approaches to long document understanding tasks.
%We demonstrate in this paper how weak supervision can be applied successfully to enable such scaling.

%% file: acknowledgments.tex
\section*{Acknowledgments}

This work wouldn’t have been possible without the detailed problem specification and data provided by UiPath. We thank them for being great collaborators and for allowing us to publish the results and dataset publicly to further the research in this critical area of computer science.